\documentclass[sigconf]{acmart}
\usepackage{times}
\usepackage{helvet} 
\usepackage{courier}  
\usepackage{graphicx} 
\usepackage{natbib}  
\usepackage{caption} 
\usepackage[utf8]{inputenc}
\usepackage{amsmath,amsthm}
\usepackage{comment}
\usepackage{algorithm,algorithmicx}
\usepackage{algpseudocode}
\usepackage{bm}
\usepackage{xcolor}
\usepackage{multirow}
\usepackage{booktabs}
\usepackage{enumerate}
\newcommand{\eat}[1]{}
\algdef{SE}[DOWHILE]{Do}{doWhile}{\algorithmicdo}[1]{\algorithmicwhile\ #1} %
{
	\theoremstyle{plain}
	
}
\eat{
\newtheorem{theorem}{Theorem}[section]
\newtheorem{lemma}[theorem]{Lemma}

}
\newenvironment{definition}[1][Definition]{\begin{trivlist}
		\item[\hskip \labelsep {\bfseries #1}]}{\end{trivlist}}

\numberwithin{equation}{section}

\setcopyright{none}
\acmISBN{}
\acmDOI{}
\acmYear{2022}
\acmConference{Conference submission}{Jan}{2022}

\DeclareMathDelimiter{(}{\mathopen} {operators}{"28}{largesymbols}{"00}
\DeclareMathDelimiter{)}{\mathclose}{operators}{"29}{largesymbols}{"01}

\begin{document}
\title[Individual Treatment Effect Estimation Through Controlled Neural Network Training]{Individual Treatment Effect Estimation Through Controlled Neural Network Training in Two Stages}
\author{Naveen Nair}
\affiliation{%
  \institution{Amazon}
  \city{Seattle}
  \country{USA}}
\email{nnair@amazon.com}

\author{Karthik S. Gurumoorthy}
\affiliation{%
  \institution{Amazon}
  \city{Bangalore}
  \country{India}}
\email{gurumoor@amazon.com}

\author{Dinesh Mandalapu}
\affiliation{%
  \institution{Amazon}
  \city{Seattle}
  \country{USA}}
\email{mandalap@amazon.com}
\eat{
\author{Anonymous}
\affiliation{%
  \institution{Anonymous institution}
}
}
\begin{abstract}
We develop a Causal-Deep Neural Network (CDNN) model trained in two stages to infer causal impact estimates at an individual unit level. Using only the pre-treatment features in stage 1 in the absence of any treatment information, we learn an encoding for the covariates that best represents the outcome. In the $2^{nd}$ stage we further seek to predict the unexplained outcome from stage 1, by introducing the treatment indicator variables alongside the encoded covariates. We prove that even without explicitly computing the treatment residual, our method still satisfies the desirable local Neyman orthogonality, making it robust to small perturbations in the nuisance parameters. Furthermore, by establishing connections with the representation learning approaches, we create a framework from which multiple variants of our algorithm can be derived. We perform initial experiments on the publicly available data sets to compare these variants and get guidance in selecting the best variant of our CDNN method. On evaluating CDNN against the state-of-the-art approaches on three benchmarking datasets, we observe that CDNN is highly competitive and often yields the most accurate individual treatment effect estimates. We highlight the strong merits of CDNN in terms of its extensibility to multiple use cases.
\end{abstract}
\maketitle

\section{Introduction}\label{sec:introduction}
Randomized trials are considered as gold standards for estimating the causal effects of treatments, by ensuring that the internal characteristics of the group who have been exposed to the treatment ---henceforth referred to as the treatment group--- and those who are not ---called the control group--- do not confound with the observed outcome. Randomization assures that any apparent difference on the outcome between these groups can be \emph{solely} attributed to the effect of the treatment. Randomization enables unbiased estimation of the treatment effect which can be directly determined by comparing the outcomes between the treated and control group. However, achieving truly random treatment and control group is infeasible and in fact unethical on many occasions. For instance one cannot choose subjects at random and request them to smoke in order to determine the effect of smoking. Hence in most of the scenarios, we only have observational (non-randomized) data where certain portion of the population undergo the treatment and some do not. 

In this work, we consider the general problem of estimating causal effects from such observational data. The objective is to determine the effect of a treatment $T$ (say drug administered to a patient) on a quantity of interest known as the outcome $Y$ (for e.g. recovery status), by controlling for subject characteristics $X$ called the covariates (such as diet, past illness, socioeconomic status). The observational data includes the subject characteristics, the treatment status and the observed outcome, but is oblivious to the process that determined the treatment status. It is often the case that the treatment group is influenced by $X$, causing systematic difference in the distribution of treatment and control populations known as selection bias or covariate shifts \cite{learningRepresentations, covariateShiftScholkopf, covariateShiftMIT}. This could lead to \emph{confounding} effects as $X$ might also directly affect $Y$.\eat{: for instance, rich people  in generally better health condition are able to afford certain medications.} Hence directly comparing the treatment and control outcomes to obtain treatment effects leads to biased estimation. The challenge in causal analysis is to eliminate such confounding factors and nullify the treatment assignment bias by properly controlling for $X$.

In this paper, we propose a two-stage Deep Learning model to minimize the selection bias in causal analysis and determine the effect of $T$ on $Y$ for a subject with characteristics $X=x$, referred to as the \emph{Individual Treatment Effect} (ITE). ITE involves estimating the causal effect of treatment at the fine grained individual level, which is much more challenging compared to estimating average treatment effects (ATE) either on the general population or average treatment effect on the treated set (ATT). In the first stage we predict the outcome $Y$ using $X$ alone, withholding the treatment status variables to obtain an encoding for $X$ that best represents the outcome. We then learn the parameters of the second stage model by introducing the treatment indicator variables alongside the encoded features and predict the unexplained outcome from stage 1. Once we have updated these parameters it is a simple exercise to obtain the individual treatment effect $\theta(x)$ for any features $X=x$. Only the second model need to be evaluated twice, once for each of the treatment and control setting, and the difference between the two predictions gives ITE. We discuss the similarities and differences of our algorithm with the Double Machine Learning approach~\cite{doubleMLdebiased} that is based on the Robinson transformation~\cite{RobinsonTransformation} and also with techniques that learn representations for causal inference~\cite{blrJohansson}, \cite{dnnShalit}, \cite{Dragonnet}. Viewing our method through the lens of the latter, we present a framework to create multiple variants of our algorithm by extracting the feature-encoding from different points in the first stage and introducing this encoding precisely at the same location in the second stage. We henceforth refer to our approach as Causal-DNN (CDNN). We compare CDNN with the state-of-the-art approaches in Section~\ref{sec:results} on three publicly available well known data sets frequently used to benchmark causal impact estimation, and demonstrate that our method is highly accurate in predicting the individual treatment effects. The paper is organized as follows. 

\eat{Our algorithm has several advantages over conventional algorithms like OLS, Propensity matching, and other state-of-the-art algorithms in terms of (i) controlling selection bias, (ii) improving scalability for multiple treatments simultaneously, (iii) computing incremental effect on multiple outcome variables simultaneously, (iv) computing heterogeneous treatment effect, (v) computing impact of increased participation of an event, (vi) nth participation impact, (vii) interaction effects of events {\it etc.} as described in Section~\ref{sec:usecases}.}

We start with a brief introduction to the notations and the assumptions behind our theory in Section \ref{sec:notations}. In Section \ref{sec:method} we present the CDNN framework, its similarity and differences with DML, satisfaction of the desirable local Neyman orthogonality condition, and connection with representation learning. A brief overview of related work is discussed in Section \ref{sec:related}. We present our results in Section \ref{sec:results} and conclude the paper in Section \ref{sec:conclusion}.

\section{Notations and assumptions} 
\label{sec:notations}
We consider the set up consisting of $N$ units indexed by $i = 1,2,\ldots, N$. Let $T_i \in \{0,1\}$ be the binary treatment indicator, with $T_i=1$ denoting that the $i^{th}$ unit has received treatment and $T_i=0$ indicating that the $i^{th}$ unit is part of the control set. We assume the existence of a pair of potential outcomes $(Y_i(1), Y_i(0))$  following Rubin causal model~\cite{potentialOutcome}. The observed outcome $Y_i$ is the potential outcome corresponding to the treatment received, namely $Y_i = T_i Y_i(1)+(1-T_i)Y_i(0)$. Let $X_i \in \mathbb{R}^d$ be the vector of covariates for the $i^{th}$ unit. Our empirical data consists of the sets of triplet: $\mathcal{D} = \{\left(Y_i, T_i, X_i\right)\}_{i=1}^N$, where each triplet is an $i.i.d.$ sample from a large population. The Individual Treatment Effect, also known as the Conditional Average Treatment Effect (CATE), which is the effect of the treatment $T_i$ on the outcome $Y_i$ for the feature $X_i=x$ is given by:
\begin{equation}
\label{eq:conditionalAverageTreatmentEffect}
\theta(x) = \mathbb{E}[Y_i(1) - Y_i(0)| X_i=x],
\end{equation}
whose expected value is the average treatment effect: $\theta = \mathbb{E}_{X}[\theta(X)]$. Empirically, the ITE for the unit $i$ is estimated as: $\hat{\theta}_i \equiv \hat{Y}_i(1) - \hat{Y}_i(0)$. We work under the standard \emph{strongly-ignorable} assumption~\cite{rubin1983}, namely $Y(0), Y(1) \perp T | X=x$ and  $0 < p(T=1|X=x) < 1$. The first condition asserts that  given $X$, the individual distribution of both $Y(0)$ and $Y(1)$ are independent of the treatment status $T=1$ or $T=0$. We also make the common simplifying assumption of \emph{no-hidden confounding}, by presuming all the factors $X$ that could potentially influence $T$ and $Y$ are observed. 
\eat{
In traditional causal models, researchers relied on linear models that assumed $f(.,x)$ to be a weighted linear combination of covariates $x$, i.e. $\mathbb{E}\left[Y(T) | x\right]  =  \alpha + \langle x , \bm{\beta}_T\rangle$. In order to mitigate the ill-effects of strong linear assumption, more sophisticated models like propensity score matching \cite{propensity1} was introduced prior to training linear models.  This enabled better causal estimation within each propensity bin that is composed of treatment and control units with similar propensity scores. Our approach explained in the subsequent section completes eliminates the need for any explicit propensity matching.
}

\section{Framework}
\label{sec:method}
Given a target feature $x \in \mathbb{R}^d$, our aim is to determine the corresponding ITE $\theta(x)$. Let $Y = f(T,x) + \epsilon$, where $\epsilon \sim \mathcal{N}(0, \sigma)$ is an independent, zero-mean Gaussian noise added to the function $f(.)$ that models the effect of $x$ and $T$ on the outcome $Y$. We then have the corresponding expected values: $\mathbb{E}\left[Y(1) | x\right] = f(T=1, x)$, $\mathbb{E}\left[Y(0) | x\right] = f(T=0, x)$ and the ITE is given by: $\theta(x) = f(T=1, x) - f(T=0, x)$. It is well known that performing a direct regression of $Y$ on $T$ and $x$ to learn the function $f(.)$ and using it to compute $\theta(x)$ is bound to produce biased estimation, because of a possible covariate shift between the treatment and control populations \cite{learningRepresentations, covariateShiftScholkopf, covariateShiftMIT}. \eat{There are several approaches that first try to match units in different population segments and compute the conditional expectation over each matched segment~\cite{rubin1983}, \cite{Austin11}. However, most of the matching is based on projections (propensity or other distance metrics) and the assumptions behind these projections may not always hold true \cite{King16}.}

In order to motivate our approach, we define $\bar{Y}$ as the conditional expectation of $Y$ given $x$ in the absence of treatment variable, i.e., $g(x) = \bar{Y} \equiv \mathbb{E}\left[Y | x\right]$ and rewrite eq.~(\ref{eq:conditionalAverageTreatmentEffect}) as $$\theta(x) = \mathbb{E}\left[(Y(1) - \bar{Y}) - (Y(0) - \bar{Y})  | X=x\right].$$ This reformulation encourages us to define a function:
\begin{equation}
\label{eq:f}
h(T,x) = f(T,x) - g(x) = \mathbb{E}\left[(Y(T) - \bar{Y}) | X=x \right]
\end{equation}
and estimate ITE as the difference between the treatment and control estimates given by:
\begin{equation}
\label{eq:ITE}
\theta(x) = h(T=1,x) - h(T=0,x).
\end{equation}
\eat{We demonstrate below that such a reformulation can indeed produce an unbiased estimate for ITE. }As explained below in Section~\ref{subsec:DML}, approaches such as Double Machine Learning (DML) \cite{doubleML} work in a similar fashion. However, the original formulation of DML is only designed for estimating ATE and does not involve conditioning on the features $x$ while regressing the residualized outcome on the residualized treatment. In order to compute ITE, DML requires parametric assumptions on $\theta(x)$ which is generally unknown \cite{chernozhukovsemiparametric}.

Our algorithm to compute $\theta(x)$ involves the 4 simple steps outlined in Algo.~\ref{algo:CDNN}. In step 1, we determine the function $g(x)$ by training a model $\mathcal{M}_1$ that takes only $x$ as input (knowledge of $T$ is deliberately suppressed), and estimates $\mathbb{E}[Y|X=x]$. The hidden layers of $\mathcal{M}_1$ compute a non-linear function of $x$ that captures most of the variance in the data in the absence of treatment indicators.\eat{We tune $\mathcal{M}_1$ on out-of-sample validation data to make sure we are not over-fitting to noises. The tuning involves selection of various parameters such as the number of hidden layers, hidden units, choice of activation functions at hidden \& output layers, learning rate, learning rate decay, number of iterations, batch size, regularization parameter, optimization function, etc.} In step 2 we compute the outcome residual $R = Y - \mathbb{E}[Y|X=x]$, where the effect of $x$ on $Y$ is partialed out. Though the outcome residual is orthogonal to features, $X$ and $T$ are generally not independent due to the presence of features that affect the treatment assignment. Hence the function $h(.)$ computed in step 3 depends on both $X$ and $T$, allowing the interaction between these two inputs, on which we regress the residual outcome. The need for including $X$ in step 3 is further emphasized in Section~\ref{subsec:DML}. As the target is changed from the actual outcome $Y$ to the residual outcome $R$, the network weights determined from $\mathcal{M}_1$ will be less useful and hence we re-initialize all the network weights to random low values. We refer to the second stage model as $\mathcal{M}_2$. The output layer in $\mathcal{M}_2$ estimates $\mathbb{E}[R|X=x, T=t]$. In step 4, we evaluate $\mathcal{M}_2$ twice for $T \in \{0,1\}$ to predict $h(T=0,x)$ and $h(T=1,x), \forall X=x$ and compute the ITE by taking the difference. Such a simple and elegant approach makes our algorithm scalable for computing ITE estimates for millions of units, as we do not need to build local regression model around every target point $x$ like in \cite{OrthogonalRandomForest} for local nuisance estimation. Scalability is required in scenarios such as estimating/forecasting the impact of a disease outbreak~\cite{causalityEpidemic}, impact of socio-economic events on citizens, etc.

\eat{ a scenario encountered routinely in large scale  e-commerce companies where causal effects are computed for different treatments like product purchases, suppression of cash on delivery, product damages due to poor packaging etc. on customer's future engagement with the company.}

\begin{algorithm}[t]
 \caption{CDNN algorithm for estimating ITE}
 \label{algo:CDNN}
\begin{algorithmic}[1]
\State Determine $g(x) = \mathbb{E}\left[Y | x\right]$ by regressing the observed outcome $Y$ on $x$ without using $T$.
\State Compute the outcome residual $Y_i-g(x_i), \forall i$.
\State Regress the outcome residual on $x$ and $T$ to learn the function $h(T,x)$.
\State For any feature $x$, estimate ITE as: $\theta(x) = h(T=1,x) - h(T=0,x)$.
\end{algorithmic}
\end{algorithm}

\subsection{Similarity and differences with DML}
\label{subsec:DML}
Let $e(x) = p\left(T=1 | X=x\right)$ be the conditional treatment probability, widely known as the \emph{propensity score}. Based on the definition of the function $g(.)$ we observe that,
\begin{align}
g(x) &= \int Y p(Y|x) \, dY = \int Y \sum\limits_{T \in \{0,1\}} p(Y,T|x) \, dY \nonumber \\
&= e(x) \int Y p(Y|x,T=1) \,dY \nonumber \\
&+ \left(1-e(x)\right)  \int Y p(Y|x,T=0) \,dY \nonumber \\
\label{eq:g}
&=e(x) f(T=1, x) + \left(1-e(x)\right)  f(T=0, x).
\end{align}
A similar form of this expression is used to derive Robinson transformation \cite{RobinsonTransformation}. This expression also leads to the following lemma. Proofs are available in the supplement.
\begin{lemma}
\label{lemma:hthetaconnection}
The function $h(T,x) = f(T,x)-g(x)$ satisfies the equality:
\begin{equation}
\label{eq:hthetae}
h(T,x) = \theta(x) \left[T-e(x)\right].
\end{equation}
\end{lemma}
\eat{
\begin{proof}
Substituting the definition of $h(T,x)$ from eq.~(\ref{eq:f}) in eq.(\ref{eq:g}), we find
\begin{align}
g(x) &= e(x) \left[h(T=1,x)+g(x)\right] + \left(1-e(x)\right) \left[h(T=0, x)+ g(x)\right], \nonumber \\
\label{eq:eheq0}
\implies &e(x) \left[h(T=1,x) - h(T=0,x)\right] + h(T=0,x) = 0.
\end{align}
Plugging the identity $\theta(x) = h(T=1, x) - h(T=0, x)$ in eq.(\ref{eq:eheq0}), we get $h(T=0,x) = \theta(x) \left[0-e(x)\right]$ and $h(T=1,x) = \theta(x) \left[1-e(x)\right]$  and the proof follows.
\end{proof}
}
When we substitute eqs.(\ref{eq:f}) and (\ref{eq:hthetae}) into the representation equation for $Y$ we get 
\begin{equation}
\label{eq:DMLsimilarity}
Y - g(x) =  \theta(x) \left[T-e(x)\right] + \epsilon.
\end{equation}
DML techniques \cite{doubleMLdebiased}, \cite{doubleML} derived using Robinson transformation advocate using eq.~(\ref{eq:DMLsimilarity}) to regress the residualized outcome $R(x)$ on the residualized treatment $\left(T-e(x)\right)$ to compute $\theta(x)$. Our step 1, where we regress $Y$ on $x$ without using the treatment indicator $T$ to determine $g(x) =  \mathbb{E}\left[Y | x\right]$ and estimate the outcome residual $Y-g(x)$ is identical to the first (machine learning) step in DML. However, the similarity ends here. The second (machine learning) step in DML involves building a propensity model to compute $e(x)$ which is then used to determine the treatment residual $T-e(x)$. The (average) treatment effect is obtained by performing regression on these orthogonalized values using the closed form expression derived in \cite{doubleML}. Extension of this approach to compute ITE requires a parametric specification of $\theta(x)$ \cite{chernozhukovsemiparametric} which is not always known. The advantage of CDNN is highlighted in this context. Our method circumvents the need to learn the propensity function and importantly, doesn't need the functional form of $\theta$ w.r.t. $x$ in order to determine ITE. Instead, the outcome residual is regressed on $x$ and $T$ to learn $h(T,x)$ and Lemma~\ref{lemma:hthetaconnection} and eq.~(\ref{eq:ITE}) is employed to directly obtain $\theta(x)$. Though the outcome residual is orthogonal to $X$, including $X$ in the step 3 of our algorithm and allowing it to interact with $T$ \emph{implicitly} determines the treatment residual. It is implicit, as in CDNN the treatment residual $T-e(x)$ is never explicitly obtained. This implicit computation avoids building a separate propensity model.

\subsection{Local Neyman orthogonality}
As explained in \cite{doubleML, OrthogonalRandomForest}, any estimate that could be proven to Neyman orthogonal is very much desirable. In such a case the ITE has \emph{reduced sensitivity} with respect to other high dimensional variables, referred to as the nuisance parameters, that directly affects its computation. Local Neyman orthogonality condition asserts that the estimate $\theta(x)$ still remains \emph{valid} under local mistakes in the nuisance variables \emph{for all} values of $x$. Mathematically it is defined using the Gateaux derivative, which is the directional derivative of the estimate w.r.t. the nuisance parameter along the direction of its perturbation, and then subsequently showing that this directional derivative is zero.

In order to establish that our solution for $\theta(x)$ satisfies local Neyman orthogonality, we need to formulate ITE in the language of conditional moment models. Given $x$, the objective is to determine the solution $\theta_0(x)$ that satisfies a system of local moment conditions \cite{OrthogonalRandomForest}, namely
\begin{equation}
\label{eq:momentcondition}
\mathbb{E}\left[\psi\left(W,\theta,\eta_0(x)\right) | X=x\right] = 0,
\end{equation}
where $\theta(x) = \theta_0(x)$ is the unique solution. Here $\psi$ is called the score function, $W = (Y, T, x)$ is the observation, and $\eta(x) = \eta_0(x)$ is the unknown, true nuisance function. The local orthogonality condition can be considered as the localized version of the Neyman orthogonality condition around the vicinity of $x$, and states that the score function $\psi$ is insensitive to local perturbations in the nuisance parameters around their true values computed at the actual ITE value $\theta_0(x)$. It is defined formally in \cite{OrthogonalRandomForest} and stated here for completeness.
\begin{definition}
Fix any estimator $\hat{\eta}$ for the nuisance function. Then the Gateaux derivative is defined as:
\begin{equation*}
D_{\psi} \left[\hat{\eta}-\eta_0 | x \right] = \mathbb{E}\left[\nabla_{\eta}\psi\left(W,\theta_0(x),\eta_0(x)\right) \left[\hat{\eta}(x)-\eta_0(x)\right]|x\right],
\end{equation*}
where $\nabla_{\eta}$ is the gradient w.r.t. $\eta$. The moment conditions are called to be \emph{locally orthogonal} if for all $x: D_{\psi} \left[\hat{\eta}-\eta_0 | x \right]=0$.
\end{definition}

As computing $\hat{\theta}(x)$ requires the knowledge of the functions $g(.)$ and $h(.)$, the nuisance parameters in our setting are given by: $\eta(x) = \left[g(x), h(.,x)\right]$. However, recall from Lemma~\ref{lemma:hthetaconnection} that the function $h(.)$ can be equivalently expressed as: $h(T,x) = \theta(x) [T-e(x)]$. At the true solution $\theta_0(x)$, any perturbation to $h(.)$ corresponds to an equivalent perturbation to the \emph{latent} function $e(x)$. We refer to the propensity function $e(x)$ as a latent variable because it is not explicitly determined in our method. Hence, the perturbations to the nuisance parameters at $\theta_0$ can be represented as: $\hat{\eta} - \eta_0 = \left[\hat{g} - g_0, \hat{e} - e_0\right]$, where $g_0(x)$ and $e_0(x)$ are the true (unknown) functions. Based on the representation of $Y$ in eq.(\ref{eq:DMLsimilarity}) where again we have used Lemma~\ref{lemma:hthetaconnection} to express $h(.)$, the score function $\psi$ that minimizes the loss $\mathbb{E}_Y \left[ \left(Y-g(x) - \theta(x)[T-e(x)]\right)^2 \mid X=x \right]$ is given by:
\begin{equation}
\label{eq:scorefunction}
\psi\left(W,\theta,\eta(x)\right) = \left(Y - g(x) - \theta(x)[T-e(x)]\right) \left(T-e(x)\right).
\end{equation}
We have the following theorem for our CDNN framework.
\begin{theorem}
For the score function $\psi(.)$ in eq.(\ref{eq:scorefunction}), the moment condition in eq.(\ref{eq:momentcondition}) respects local Neyman orthogonality. 
\end{theorem}
Our insight into the functional form of $h(.)$ in Lemma~\ref{lemma:hthetaconnection} is essential for this technical property to hold true. \emph{We are not aware of any prior work that establishes local orthogonality without explicitly computing the treatment residual.}

\subsection{Connection with representation learning}
\label{subsec:representationLearning}
\begin{figure*}
\centering
	\includegraphics[scale=0.4]{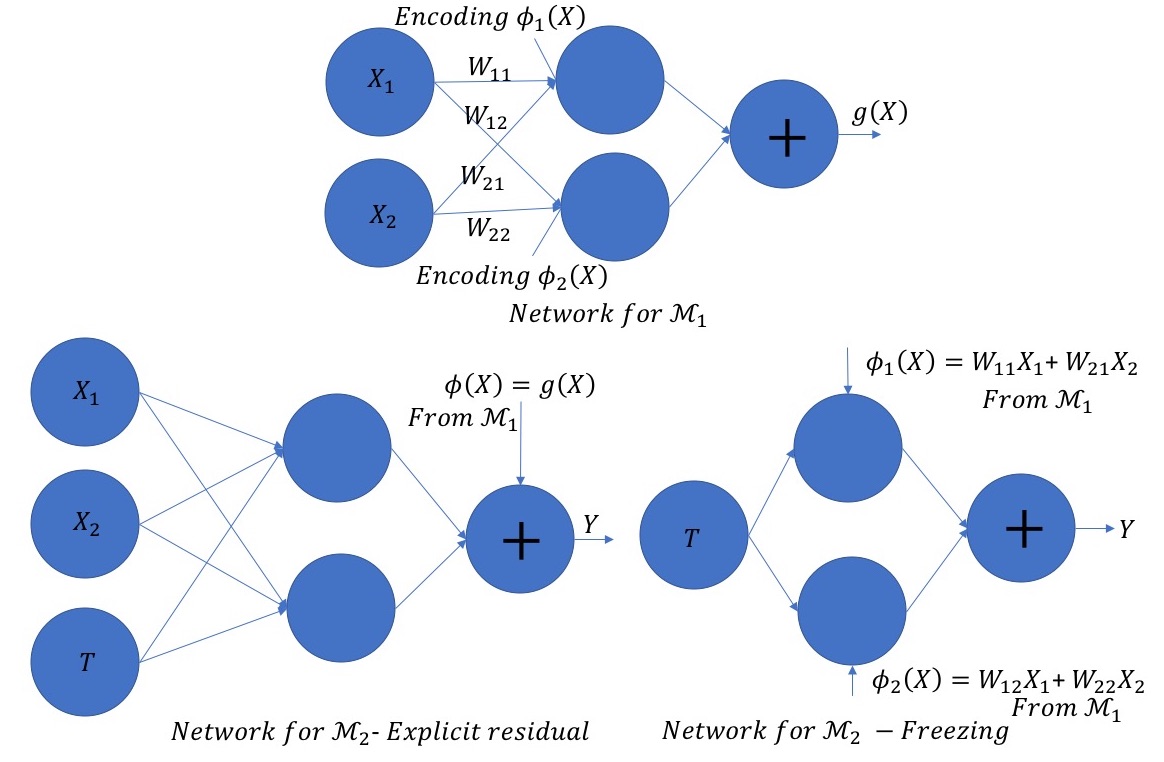}
	\caption{CDNN network structure (depicts $2-$dimensional $X$ and $1-$hidden layer only for illustration) for \emph{explicit-residual} and  \emph{freezing} approaches. Encoding $\phi(x)$ is computed in $\mathcal{M}_1$ and employed precisely at the same point in $\mathcal{M}_2$.}
	\label{fig:CDNNencoding}
\end{figure*}

There has been numerous research works that try to learn latent representations for the features, with the aim of accurately estimating the treatment effects \cite{blrJohansson}, \cite{dnnShalit}, \cite{Dragonnet}. The underlying principle is to find an encoding for $x$, denoted as $\phi(x)$, using which a loss function $L\left(h(T,\phi(x)),Y\right)$ is minimized by regressing the outcome $Y$ on $\phi(x)$ and $T$ to learn a function $h(.)$ which is used to compute ITE via: $\theta(x) = h(T=1,\phi(x)) - h(T=0,\phi(x))$. For instance, the CFR algorithm in~\cite{dnnShalit} finds that encoding $\phi(x)$ where treatment and the control population are balanced. The Dragonnet method in \cite{Dragonnet} produces the representation $\phi(x)$ that distills the covariates into the features relevant for propensity score estimation. Our residual based CDNN method, though on the face of it appears to have little resemblance with representation learning, indeed is suitably producing an encoding for $x$. To see this, recall that in $\mathcal{M}_1$ we regress $Y$ on $x$ to compute $g(x)$, and the outcome residual $Y-g(x)$. This value $g(x)$ can be interpreted as the one-dimensional encoding of $x$ ($\phi(x)$), capable of representing $Y$ in the best possible manner. Similar to CFR method \cite{dnnShalit} that sets $\phi(x)$ to be that representation where treatment and control populations are balanced, like the Dragonnet \cite{Dragonnet} where $\phi(x)$ is the transformation of $x$ germane for treatment prediction, CDNN encodes $x$ into that one-dimensional $\phi(x)$ that best explains $Y$. Like the other two methods $\phi(x)=g(x)$ is used in $\mathcal{M}_2$, but CDNN uses it precisely at the same point where this encoding is extracted from $\mathcal{M}_1$. Since $\phi(x)$ is the representation of $x$ at the output layer in $\mathcal{M}_1$, $\phi(x)$ should feature only at the output layer in $\mathcal{M}_2$, i.e., added to the function $h(T,x)$ in minimizing the loss: $L\left(h(T,x)+\phi(x),Y\right)$. This process is visualized in Fig.~(\ref{fig:CDNNencoding}). In the special case where this loss function is a squared loss, it becomes equivalent to calculating the residue $Y-g(x)$ and learning the function $h(.)$ that minimizes $\|h(T,x) - (Y-g(x))\|^2$. Note that $h(.)$ is also a function of $x$ as the encoding $\phi(x)$ does not play an active role in learning $\mathcal{M}_2$, and merely shows up at the output. 

This key insight that CDNN is also learning a representation opens the possibility to develop multiple variants of our method. Consider the case where the encoding of $x$ is extracted at the input layer in $\mathcal{M}_1$, i.e., setting $\phi(x)=x$ with no actual encoding and employed at the input layer in $\mathcal{M}_2$. This is equivalent to modeling $Y = f(T,x) + \epsilon$ and directly regressing $Y$ on $T$ and $x$ in a single step. As stated earlier, calculating $\theta(x) = f(T=1,x) - f(T=0,x)$ is known to be biased as it does not control for covariate shifts. A useful variant could be to use the linear transformation of $x$, sent as input to the first hidden layer $\mathcal{H}^1$ in model $\mathcal{M}_1$, as $\phi(x)$ and then employ this encoding precisely at the same point in $\mathcal{M}_2$ as shown in the $3^{rd}$ sub-figure of Fig.~\ref{fig:CDNNencoding}. Such an approach, referred to as \emph{freezing} to differentiate it from the alternative that involves computing the residual $R(x)$ called \emph{explicit-residual}, comprises of the steps presented in Algo.~\ref{algo:freezing}.

\begin{algorithm}[t]
 \caption{The \emph{freezing} variant of CDNN}
 \label{algo:freezing}
\begin{algorithmic}[1]
\State In $\mathcal{M}_1$, regress the observed outcome $Y$ on $X$ without using $T$ as shown in Fig.~\ref{fig:firstStageModel}.
\State Set $\phi(x) = Wx$ as the encoding, where $W$ is the weight matrix between $x$ and $\mathcal{H}^1$.
\State Introduce $T$ as additional input to $\mathcal{M}_2$ and freeze the weight matrix $W$ (do not update) so that $\phi(x)$, instead of $x$, is the input as shown in Fig.~\ref{fig:secondStageModel}.
\State Regress $Y$ on $T$ and $\phi(x)$ to learn the function $h(T,\phi(x))$ and calculate ITE as $\theta(x) = h(T=1,\phi(x)) - h(T=0,\phi(x))$. The network weights between $T$ and $\mathcal{H}^1$ are updated in each epoch.
\end{algorithmic}
\end{algorithm}
We show the practical utility of this variant in the experimental Section~\ref{sec:results}. The advantage of the freezing variant is two-fold. Firstly, as both $\mathcal{M}_1$ and $\mathcal{M}_2$ are trained to predict the actual outcome $Y$ (unlike the explicit residual approach where outcome residual is used in $\mathcal{M}_2$), the trained weights from $\mathcal{M}_1$ between the hidden layers and between the final hidden and the output layers can be used as the initial condition in $\mathcal{M}_2$ without re-initializing them. Secondly, while training $\mathcal{M}_2$ the treatment variable parameters are updated only if there is more information in them to infer the outcome. This helps to quantify the treatment effect accurately. For instance, consider the scenario where $X$ can predict both the treatment indicator $T$ and outcome $Y$ accurately such that both the treatment residual and the outcome residual are close to zero. Since there is no evidence for any treatment effect, one could argue that ITE should also be closer to zero. It is easy to see that in such cases $h(T,\phi(x)) = 0$, for both values of $T$ resulting in an estimate of $\theta(x)=0$. In contrast, approaches such as DML could become sensitive to noise due to the possible scale difference between the treatment and outcome residuals. 

In summary the following are the strengths of our approach:
\begin{enumerate}[{(}1{)}]
\item Computes ITE via a simple difference operation without modeling the propensity function $e(x)$ to determine the treatment residual.
\item Unlike DML techniques  \cite{doubleMLdebiased}, \cite{doubleML}, our method does not require to explicitly parametrize $\theta$ in terms of $x$ for estimating ITE.
\item Implicitly satisfies the desirable local Neyman orthogonality.
\item A scalable algorithm to obtain ITE estimates for millions of units as it avoids building local regression models around every target point $x$.
\item Trivially extend-able to be used in practical situations such as multiple treatments, continuous treatments, treatment interactions and multiple outcomes. Discussion and evaluation on these are future works and are briefly outlined in section \ref{sec:conclusion}. 
\end{enumerate}

\begin{figure}
	\centering
	\begin{minipage}{.45\textwidth}
	\centering
	\includegraphics[scale=0.22]{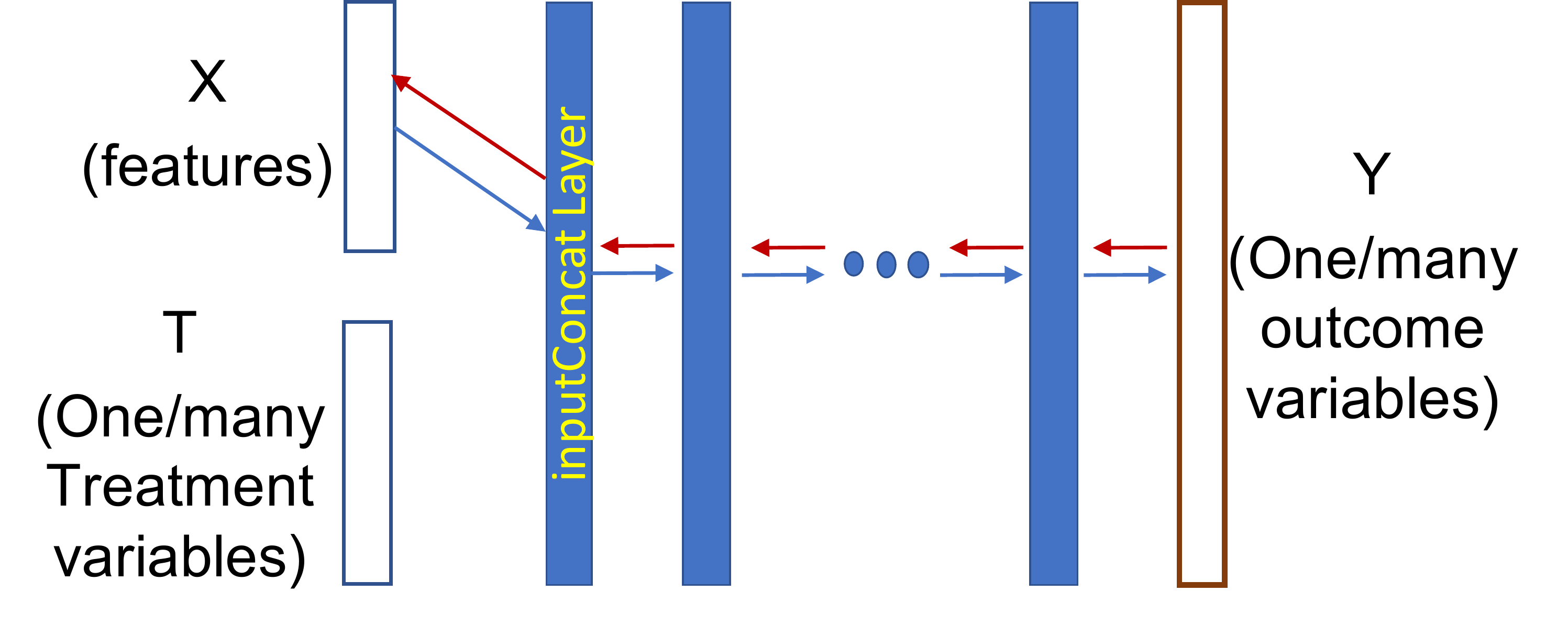}
	\caption{Stage 1. Right arrow indicates forward propagation of weights and left arrow indicates back propagation of gradients. \eat{Please note that the weights and the weight updates are for the edges connecting layers, although the arrows are drawn to end at layers. In first stage, only non-treatment inputs are used to learn the model.}}
	\label{fig:firstStageModel}
\end{minipage}%
\hfill
\begin{minipage}{.45\textwidth}
	\centering
	\includegraphics[scale=0.22]{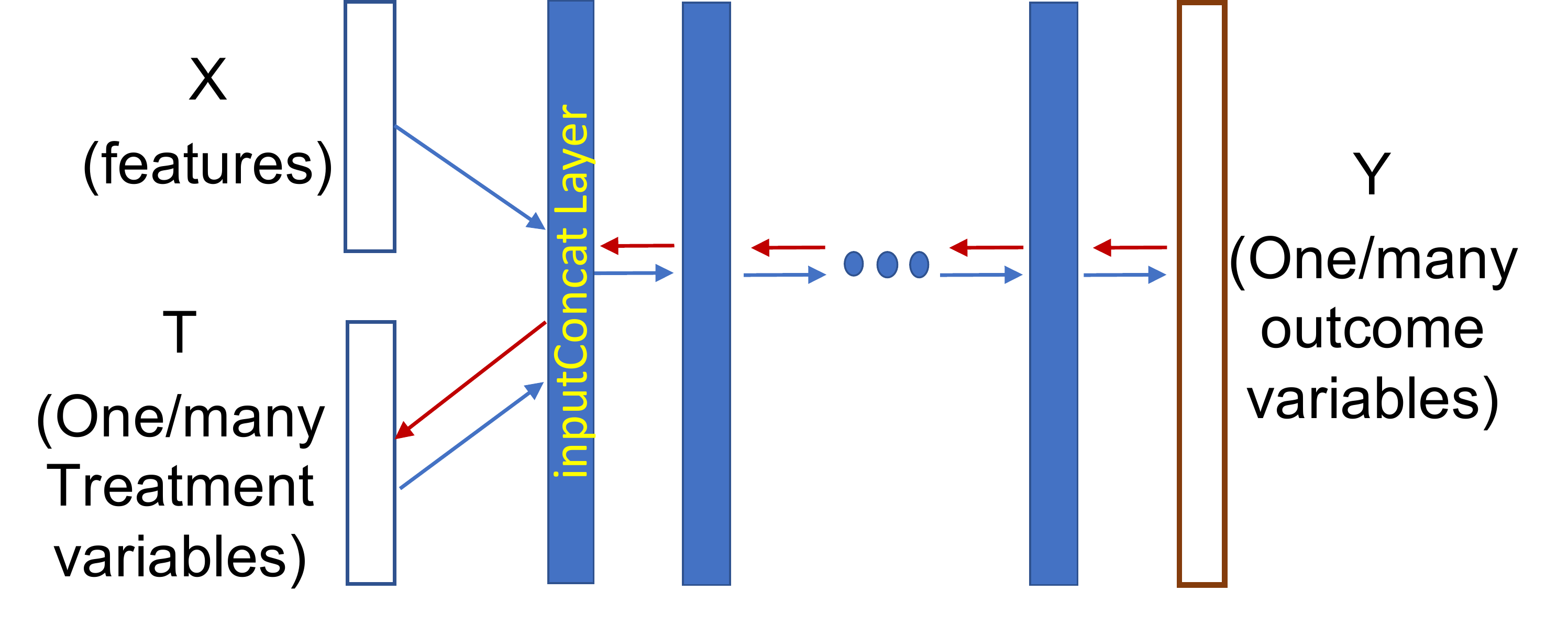}
	\caption{Stage 2. Right arrow indicates forward propagation of weights and left arrow indicates back propagation of gradients. \eat{Please note that the weights and the weight updates are for the edges connecting layers, although the arrows are drawn to end at layers. In second stage, non-treatment inputs parameters are not updated and gradients are propagated only for treatment variables.}}
	\label{fig:secondStageModel}
	\end{minipage}
\end{figure}
\section{Related work}
\label{sec:related}
In Sections~\ref{subsec:DML} and \ref{subsec:representationLearning} we described the DML and the representation learning methods and discussed their similarities and differences with CDNN. For the representation algorithm in \cite{blrJohansson} that encourages balancing the treatment and control populations, Shalit {\it et al.} \cite{dnnShalit} derived a generalization-error bound for ITE in terms of sum of standard generalization error of the representation and the distance between the induced treated and control distributions. Crump {\it et al.} \cite{knnCrump} developed non-parametric tests to identify if the treatment has a non zero average effect for any subpopulation and whether the average effect is identical for all sub-populations. K{\"u}nzel {\it et al.} \cite{XLearner} proposed a meta-learner framework named X-learner that estimates ITE by learning separate conditional outcome estimators for treatment and control populations. Farrell  {\it et al.}  \cite{JointLearnDNN} developed a deep neural network based approach for estimating ITE based on combined training of two networks. The first network is trained to predict the control outcome and the second network is trained on the residual to directly predict the ITE. Hitsch  {\it et al.} \cite{OutcomeTransformITE} suggested a transformation based method that suitably modifies the outcome for estimating ITE. A doubly robust approach for estimating ITE is given in \cite{DRITE}. XBART, a nonparametric Bayesian regression approach using trees constrained by a regularization prior to be a weak learner, is proposed by Chipman {\it et al.} \cite{bartChipman}. Founded on the work by Athey and Imbens \cite{guidoMLmethods} on using machine learning methods for estimating heterogeneous causal effects,  Wager and Athey \cite{causalForest} developed a non-parametric causal forest approach that leverages the Random Forest \cite{randomForest} algorithm and showed that causal forest gave point wise consistent treatment effect estimation. The Generalized Random Forest (GRF) approach developed in \cite{GRF} is another nonparametric method for heterogeneous treatment effect estimation based on random forests. The Orthogonal Random Forest (ORF) algorithm in \cite{OrthogonalRandomForest} is an extension of \cite{GRF}. Here the authors present a theoretical foundation for residual on residual regression approach recommended in \cite{GRF} through the lens of Neyman orthogonality, and perform residualization locally around the target estimation point $x$ as opposed to performing overall residualization like in \cite{GRF}. The heterogeneity in both GRF and ORF are obtained by deriving an adaptive weighting function from the forest and using them to weight the training examples. As different sets of training instance weights have to be learned for each estimation point $x$, these methods do not scale for estimating ITE for hundreds of thousands of samples. The work in~\cite{cmgp} proposed a multi-task learning framework that models the factual and counterfactual outcomes as outputs of a function in a vector-valued Reproducing Kernel Hilbert space (RKHS). GANITE~\cite{GANITE} is a framework for estimating the individual treatment effects by leveraging Generative Adversarial Nets to capture the uncertainty in the counter-factual distributions. 

In relation to these methods our two-stage CDNN formulation is motivated from computing \emph{incremental value} of using treatment variable to predict outcome, which intuitively aligns with the objective of estimating the incremental effect of treatment on outcome of interest. \eat{To the best of our knowledge, there is no existing work similar to the method proposed in this paper. In addition to the empirical evidence presented in the next section, we firmly believe that no other existing approach is as scalable or extendable as our CDNN method with regard to the aspects discussed in Section~\ref{sec:usecases}.}
\section{Experiments and results}
\label{sec:results}
Evaluating causal inference algorithm on real world data is a difficult task because of underlying unmeasured bias and missing counterfactual outcomes. Recall that in the real world, every unit is either the part of the treatment set or the control group but never in both. We never have both the potential outcomes $Y_i(1)$ and $Y_i(0)$ for any units $i$, and therefore the lack of ground truth individual causal effect values $\theta_i = Y_i(1)-Y_i(0)$. The best that could be done for evidence based validation of a proposed algorithm is either: (i) use a semi-synthetic data where both treatment and control outcomes are synthetically generated and true ITE values are determined or (ii) obtain ITE values under specific structure of dataset where the outcome of the other known twin pair is considered as counterfactual. Because of very limited availability of datasets with ground truths, almost all the approaches show experimental results following the same methodology. In this work we use such popularly accepted benchmarking datasets to perform our experiments.

\textbf{Implementation:} We implemented our approach in Python using mxnet-Gluon \cite{gluon} package for DNN
\footnote{Code and sample data available at github.com/KSG0507/DOEC.\eat{Every effort is taken to ensure that author names and affiliations are kept anonymous in the github link}}. 
We specified two input layers, one each for features and treatment variables, which are concatenated to form the \emph{inputConcat} layer and then sent as input to the succeeding layers. We feed the inputConcat layer to every succeeding layer as it tends to improve the outcome prediction accuracy. This is an optional step for efficiency and does not change the causal interpretation of CDNN. We configured between 3-10 fully connected hidden layers with 512-1024 nodes at each layer and used Swish as the activation function. Weights for all edges relating to treatment variables are initialized to 0 and not allowed to update weights in the first stage. The purpose of first stage model is to separate nuisance parameters from the parameters of interest and therefore the first stage prediction results does not provide much insights that would help the interpretation of causal estimates. In the second stage, the weights for treatment variables are set to small random values and allowed to update. However, we freeze weight learning for the edges corresponding to features in the second stage; but the weights are propagated in the forward pass. This ensures that the contribution of features learned in first stage is not altered in the second stage.\eat{the activation function applied on each node in the inputConcat layer helps to prevent any succeeding layer from identifying how the output from inputConcat layer nodes are computed.} 

The standard $\ell_2$ difference between the predicted and observed $Y$ was set as the loss function. To mimic the real world scenario, the actual causal effect values though known on the experimental datasets were \emph{never used} during any part of training or validation. We split train data into multiple train and validation sets and train one model for each split. The predictions from these multiple models were averaged to get the final outcome prediction (ensemble) for both stages. We trained networks with 100-2500 epochs for each stage and batch sizes ranging from 60-512. Experiments were run on a CPU that has 32 cores and 256GB RAM. Following the standard practice in causal inference works, we present standard deviation for the error metrics so that our results are directly comparable with other methods.

\textbf{Data}: We ran experiments on the three popularly used datasets ---IHDP \cite{ihdpHill}, Twins \cite{GANITE}, \cite{twinDataNBER}, and News \cite{learningRepresentations}--- having ground-truth data for true ITE values and pitted CDNN against other state-of-the-art methods. 

(1) IHDP is a semi-synthetic dataset constructed from measurements relating to children and their mothers from an experiment done in 1985. The experiment was to understand the effect of home visits by specialists on future cognitive test scores for the kids under study. There are 25 features in the data to estimate impact on one outcome variable. For using as benchmarking dataset, Hill \cite{ihdpHill} created an imbalance by removing a biased subset from treatment set. A treated and a control outcome is simulated to represent the conditional expectation function at each feature vector $x$, so that the true individual causal effect is known. \eat{The factual and counter-factual outcomes are created from the conditional treated and control outcome by adding an error term.}It consists of 747 samples, replicated 1000 times with different data generation processes and are available for download through the NPCI package~\cite{NPCI}. 

(2) Twins data is created from twin births in USA between 1989 and 1991~\cite{twinDataNBER} containing 11,400 pairs of twins with weight less than 2kg at birth. Heavier and lighter infants among the twins are assigned to treatment and control sets respectively, and the child's mortality after one year is set as binary outcome. There are 30 features derived from parents, pregnancy and birth attributes. Since the outcome for both twins are known, the true individual treatment effect is known. To simulate an observational study, only one of the two twins are included in the data with a conditional bias induced in selecting the observable twin. The simulation is replicated 100 times with different conditional bias. Please refer to \cite{GANITE}, \cite{twinDataNBER} for more details on the data. \eat{We process the data following the steps outlined in~\cite{GANITE} and request the reader to refer to it for more details.}

(3) News data introduced in \cite{learningRepresentations}, is simulated based on a topic model trained on NY Times document corpus \cite{NewsDataRepoUCI}. Each unit in the data represents a news item with word counts from a vocabulary of 3477 words as features. Treatment status is assigned based on the reading device (mobile for treatment and desktop for control) and the outcome is the experience of reading simulated from the topic model. 5000 news items and outcomes are sampled based on 50 LDA topics.

\textbf{Design:} We followed the same train-test splits adopted in the literature. We split each replication of IHDP into $63\%$ train, $27\%$ validation and $10\%$ test as in ~\cite{Dragonnet}. For each replication of the Twins and News data, we split into $56\%$ train, $24\%$ validation and $20\%$ test~\cite{GANITE, learningRepresentations}. In all the replications we made 3 separate train/validation splits to build our ensemble model. Maintaining such level of consistency on the data used for training, validation, and testing enables us to directly use the reported results in previous papers, a practise closely followed in the literature~\cite{dnnShalit, GANITE, Dragonnet}. We compare approaches based on the Precision in Estimation of Heterogeneous Effect $\left(\epsilon_{PEHE}\right)$ defined as: $$\sqrt{\epsilon_{PEHE}} = \sqrt{\frac{1}{N}\sum_{i=1}^{N}\Big(\big(\hat{Y}_i(1)-\hat{Y}_i(0)\big) - \big(Y_i(1) - Y_i(0)\big)\Big)^2}$$ as defined in~\cite{ihdpHill}, \cite{GANITE}.
 \eat{and the error in Average Treatment Effect \big($\epsilon_{ATE} = ${\scriptsize$\frac{1}{N}\sum_{i=1}^{N}\big((\hat{Y}_i(1)-\hat{Y}_i(0)) - (Y_i(1) - Y_i(0))\big)$} \Big) }

\textbf{Results:} 
As discussed in Section \ref{subsec:representationLearning} our approach can be implemented in two ways: (1) \emph{explicit residual variant} where we train the second stage using the residual as the target and (2) \emph{freezing variant} where during the training phase of the second stage, the weights associated with the features $X$ obtained from stage 1 are not allowed to update and the target is set as the original observed outcome and not the residual. While the former uses a one-dimensional encoding for $X$ in the second stage, the latter employs a high-dimensional, linear transformation of $X$ as its representation. We compare these two implementations on a smaller 10 replications each from the IHDP, Twins and News data. Table~\ref{table:explictVsFreezingPEHE} presents the  $\sqrt{\epsilon_{PEHE}}$ comparison on out-of-sample datasets. We observe that the freezing variant is able to estimate causal effects more accurately compared to the explicit residual approach with low error values. The superiority of the freezing variant is because of the following reasons:
\begin{enumerate}[{(}1{)}]
    \item It learns and preserves a higher dimensional representation of $X$ in its hidden layers that predicts the outcome.
    \item By avoiding to explicitly determine outcome residuals $Y-g(x)$, any error in the residualized outcome computation from the first stage is not transferred to the subsequent stage, as the freezing variant uses the true outcome values in both the first and second stages.
\end{enumerate}
\eat{We reason that by avoiding the need to explicitly compute residual and by preserving a high dimension representation of $X$ that predicts the outcome, the freezing method results in better treatment effect estimation. }We therefore recommend and used the freeze layer implementation for the rest of our experiments.
\begin{center}
	\begin{table}[!ht]
		\centering
		\caption{Out-of-sample $\sqrt{\epsilon_{PEHE}}$ comparison of explicit residual computation and freeze layer implementation on 10 replications each of IHDP, Twins and News data. }
		\label{table:explictVsFreezingPEHE}
		\begin{tabular}{|l|r|r|}
			\hline
			Data & Explicit Residual & Freeze Layer\\
			\hline
			IHDP &   $1.65\pm3.28$  & $\bm{0.65}\pm1.00$ \\
			Twins &   $0.32\pm0.01$  & $\bm{0.32}\pm0.01$ \\
			News &   $3.85\pm1.10$ & $\bm{1.83}\pm0.35$ \\
			\hline		
		\end{tabular}
	\end{table}
\end{center}

We compared CDNN with several approaches such as GANITE \cite{GANITE}, Ordinary Least Squares regression (OLS), least squares regression using treatment as a feature (OLS/LR$_1$), separate least squares regressions for each treatment (OLS/LR$_2$), Lasso + Ridge regression\cite{learningRepresentations}, balancing linear regression (BLR) \cite{blrJohansson}, k-NN \cite{knnCrump}, BART \cite{bartChipman}, Random Forest (RF) \cite{randomForest}, Causal Forest (CF) \cite{causalForest}, different variants of balancing neural network (BNN) like BNN-4-0, BNN-2-2 \cite{blrJohansson}, Feed-forward Neural Network with 4 hidden layers (NN) \cite{blrJohansson}, TARNET \cite{dnnShalit}, CFR$_{W ASS}$ \cite{dnnShalit}, multi-task gaussian process (CGMP) \cite{cmgp}, Double ML \cite{doubleML} and Doubly Robust regression (Doubly Robust) \cite{doublyRobust}.
Following the standard procedure, we considered 1000 replications of IHDP dataset, 100 replication of Twins and 50 replications of News Data in order to determine \eat{confidence intervals}standard deviation for our estimates \cite{GANITE}. The results on the $\sqrt{\epsilon_{PEHE}}$ metric for IHDP, Twins and News are presented in Tables~\ref{table:PEHE_IHDP}, \ref{table:PEHE_Twins} and ~\ref{table:PEHE_News} respectively. As followed by other authors \cite{dnnShalit}, \cite{GANITE}, we referred to \cite{GANITE} for IHDP and Twins, and \cite{learningRepresentations} for News to obtain the $\sqrt{\epsilon_{PEHE}}$ values for other approaches.
The following are the few important observations from our experimental results.
\begin{enumerate}[{(}1{)}]
\item  CDNN is highly competitive and provides highly accurate individual causal effect estimates. Specifically, CDNN incurs the lowest error value on two of the three datasets.
\item On the most popular IHDP dataset, the error in terms of $\sqrt{\epsilon_{PEHE}}$ metric of the second best algorithm CFR$_{WASS}$, equalling $0.76$, is $1.4$ times higher than CDNN whose error is as low as $0.54$.
\item  Likewise on the News dataset, barring one method BNN-2-2, every other algorithm has statistically much higher error value compared to CDNN.
\item The performance of all the algorithms, including CDNN, are statistically similar to each other on the Twins dataset.
\end{enumerate}
 We would like to highlight that our method takes less than a second to train per epoch. The flexible framework of CDNN could further be leveraged to compute ITE's estimates for millions of units and can be applied to many other use cases, as discussed in Section~\ref{sec:conclusion}, further adding to its strength.

\eat{
\begin{center}
	\begin{table}[!ht]
		\centering
		\caption{$\sqrt{\epsilon_{PEHE}}$ comparison of explicit residual computation and freeze layer implementation on 10 replications each of IHDP, Twins and News data. Results are on out-of-sample data.}
		\label{table:explictVsFreezingPEHE}
		\begin{tabular}{|l|l|r|r|}
			\hline
			Metric&Method &IHDP&Twins\\ 
			\hline
			\multirow{2}{*}{$\sqrt{\epsilon_{PEHE}}$}&Explicit Residual &   $1.65\pm3.28$  & $0.320\pm0.007$ \\
			&Freeze Layer &   $\bm{0.65}\pm1.00$ & $\bm{0.317}\pm0.007$ \\
			\hline	
			\multirow{3}{*}{$\epsilon_{ATE}$}&Explicit Residual & $0.16\pm0.24$&  $0.015\pm0.006$ \\
			&Freeze Layer &   $\bm{0.10}\pm0.06$& $0.007\pm0.006$ \\
			&DML  & $0.74\pm1.76$ &$\bm{0.003}\pm0.003$\\ 
			\hline		
		\end{tabular}
	\end{table}
\end{center}
}

\begin{center}
	\begin{table}[!h]
		\centering
		\caption{Out-of-sample $\sqrt{\epsilon_{PEHE}}$  results on IHDP ($1000$ replications) dataset. }
		\label{table:PEHE_IHDP}
		\begin{tabular}{|l|r|}
			\hline
			Method &$\sqrt{\epsilon_{PEHE}}$\\ 
			 \hline
			 CDNN &  $\bm{0.54}\pm0.32$ \\
			 \hline
			 GANITE &   $2.40\pm0.40$  \\
			 OLS/LR$_1$&   $5.80\pm0.30$ \\
			 OLS/LR$_2$&   $2.50\pm0.10$ \\
			 BLR&   $5.80\pm0.30$ \\
			 k-NN&   $4.10\pm0.20$ \\
			 BART  & $2.30\pm0.10$\\
			 RF&  $6.60\pm0.30$ \\
			 CF&  $3.80\pm0.20$ \\
			 BNN&  $2.10\pm0.10$ \\
			 TARNET&   $0.95\pm0.02$ \\
			 CFR$_{W ASS}$  & $0.76\pm0.02$ \\
			 CMGP& $0.77\pm0.11$ \\
			\hline			
		\end{tabular}
	\end{table}
\end{center}

\begin{center}
	\begin{table}[!h]
		\centering
		\caption{Out-of-sample $\sqrt{\epsilon_{PEHE}}$  results on Twins ($100$ replications) dataset. }
		\label{table:PEHE_Twins}
		\begin{tabular}{|l|r|}
			\hline
			Method &$\sqrt{\epsilon_{PEHE}}$\\ 
			\hline
			CDNN &    $0.319\pm0.008$ \\
			\hline
			GANITE&  $\bm{0.297}\pm0.016$\\
			OLS/LR$_1$ &   $0.318\pm0.007$\\
			OLS/LR$_2$ &  $0.320\pm0.003$\\
			BLR &     $0.323\pm0.018$\\
			K-NN &   $0.345\pm0.007$\\
			BART &   $0.338\pm0.016$\\
			RF &      $0.321\pm0.005$\\
			CF &      $0.316\pm0.011$\\
			BNN &       $0.321\pm0.018$ \\
			TARNET&       $0.315\pm0.003$\\
			CFR$_{W ASS}$&    $0.313\pm0.008$\\
			CMGP &        $0.319\pm0.008$ \\
			\hline			
		\end{tabular}
	\end{table}
\end{center}

\begin{center}
	\begin{table}[!h]
		\centering
		\caption{Out-of-sample $\sqrt{\epsilon_{PEHE}}$  results on News ($50$ replications) dataset. }
		\label{table:PEHE_News}
		\begin{tabular}{|l|r|}
			\hline
			Method  &$\sqrt{\epsilon_{PEHE}}$\\ 
			\hline
			CDNN   &  $\bm{1.9}\pm0.4$  \\
			\hline
			OLS &   $3.3\pm0.2$   \\
			Doubly Robust   &   $3.3\pm0.2$ \\
			Lasso + Ridge  &  $3.4\pm0.2$  \\
			BLR &    $3.3\pm0.2$  \\
			BNN-4-0   &  $3.4\pm0.2$  \\
			NN-4 &   $3.8\pm0.2$ \\
			BART &   $3.2\pm0.2$  \\
			BNN-2-2 & $2.0\pm0.1$  \\
			\hline			
		\end{tabular}
	\end{table}
\end{center}

\subsection{Computing average treatment effects}
Estimating ITE is much more challenging compared to computing average treatment effects like ATE or ATT, as it involves determining causal values at a fine grained individual level. However once ITE values are obtained, estimating ATE or ATT is straightforward requiring simple average computation on the appropriate population. The reverse is not true and hence techniques like Dragonnet~\cite{Dragonnet}, DML~\cite{doubleML} which are primarily designed to determine ATE cannot be employed to obtain ITE. We ran CDNN to estimate ATE and compared against other approaches based on the following error metric: $$\epsilon_{ATE} = \frac{1}{N}\sum_{i=1}^{N}\Big((\hat{Y}_i(1)-\hat{Y}_i(0)\Big) - \Big(Y_i(1) - Y_i(0)\Big)$$ where $\Big(\hat{Y}_i(1), \hat{Y}_i(0)\Big)$ are the predicted outcomes under the treatment and control settings respectively, and $\Big(Y_i(1), Y_i(0)\Big)$ are the available ground truth values. It is worth emphasizing that lower the value of this error metric, better is the accuracy of the method in predicting the average treatment effects. The results for IHDP, Twins and News datasets are shown in Tables~\ref{table:ATE_PEHE_IHDP}, \ref{table:ATE_PEHE_Twins} and \ref{table:ATE_PEHE_News} respectively. While CDNN outperform other approaches on the IHDP data, it is definitely competitive on other two data sets. \emph{We observe that no other algorithm has consistently low $\epsilon_{ATE}$ values like CDNN on all the three datasets.}

\begin{center}
	\begin{table}[h]
		\centering
		\caption{$\epsilon_{ATE}$ comparison on IHDP ($1000$ replications) dataset. Results are on out-of-sample data except for DML that has results on entire data}
		\label{table:ATE_PEHE_IHDP}
		\begin{tabular}{|l|r|}
			\hline
			Method &$\epsilon_{ATE}$ \\ 
			\hline
			CDNN & $\bm{0.13}\pm0.10$\\
			\hline
			GANITE &  $0.49\pm0.05$\\
			OLS/LR$_1$&  $0.94\pm0.06$\\
			OLS/LR$_2$&  $0.31\pm0.02$\\
			BLR&  $0.93\pm0.05$ \\
			k-NN&   $0.90\pm0.05$ \\
			BART& $0.34\pm0.02$\\
			RF& $0.96\pm0.06$ \\
			CF& $0.40\pm0.03$ \\
			BNN& $0.42\pm0.03$ \\
			TARNET&  $0.28\pm0.01$ \\
			CFR$_{W ASS}$& $0.27\pm0.01$ \\
			CMGP& $\bm{0.13}\pm0.12$\\
			CEVAEs&  $0.46\pm0.02$\\
			Dragonnet &  $0.21\pm0.01$ \\
			Dragonnet + t-reg&  $0.20\pm0.01$\\
			DML  & $0.69\pm1.13$\\ 
			\hline			
		\end{tabular}
	\end{table}
\end{center}

\begin{center}
	\begin{table}[h]
		\centering
		\caption{$\epsilon_{ATE}$ comparison on Twins ($100$ replications) dataset. Results are on out-of-sample data except for DML that has results on entire data}
		\label{table:ATE_PEHE_Twins}
		\begin{tabular}{|l|r|r|}
			\hline
			Method &$\epsilon_{ATE}$\\ 
			\hline
			CDNN &   $0.006\pm0.005$ \\
			\hline
			GANITE&  $0.009\pm0.008$\\
			OLS/LR$_1$ &   $0.007\pm0.006$\\
			OLS/LR$_2$ &   $0.007\pm0.006$\\
			BLR &   $0.033\pm0.009$\\
			K-NN &    $0.005\pm0.004$\\
			BART &   $0.127\pm0.023$\\
			RF &    $0.008\pm0.005$\\
			CF &    $0.034\pm0.008$\\
			BNN &      $0.020\pm0.007$\\
			TARNET&    $0.015\pm0.002$\\
			CFR$_{W ASS}$&   $0.028\pm0.003$\\
			CMGP &     $0.014\pm0.012$\\
			DML  &$\bm{0.004}\pm0.003$\\ 
			\hline			
		\end{tabular}
	\end{table}
\end{center}

\begin{center}
	\begin{table}[h]
		\centering
		\caption{$\epsilon_{ATE}$ comparison on News ($50$ replications) dataset. Results are on out-of-sample data except for DML that has results on entire data}
		\label{table:ATE_PEHE_News}
		\begin{tabular}{|l|r|r|}
			\hline
			Method &$\epsilon_{ATE}$\\ 
			\hline
			CDNN &  $0.3\pm0.2$ \\
			\hline
			OLS &  $\bm{0.2}\pm0.0$ \\
			Doubly Robust &  $\bm{0.2}\pm0.0$ \\
			Lasso + Ridge &  $0.6\pm0.0$ \\
			BLR &  $0.6\pm0.0$ \\
			BNN-4-0 &  $0.3\pm0.0$ \\
			NN-4 &  $1.1\pm0.0$\\
			BART & $\bm{0.2}\pm0.0$ \\
			BNN-2-2 & $0.3\pm0.0$ \\
			DML  & $0.4\pm0.3$\\ 
			\hline			
		\end{tabular}
	\end{table}
\end{center}

\section{Conclusion}
\label{sec:conclusion}
In this paper we proposed a two-stage DNN based causal impact estimation method that first learns a representation for the features relevant for outcome prediction, and then subsequently adds the treatment variables to determine the incremental effect of treatment. We formally derived our approach and established connection with DML and representation learning approaches. Our empirical evaluation on benchmarking datasets shows that CDNN is highly competitive and outperforms most state-of-the-art approaches in estimating treatment effects.

As part of our future work, following are some of the possible extensions and applications of the CDNN framework.\\ (1) \emph{Multiple Treatments:} Although we have presented CDNN from the viewpoint of a single binary treatment, it is readily extensible to the multiple treatment scenario. Typical causal models tend to run separate models for each treatment by keeping other treatment variables as control variables, which is computationally expensive. However in CDNN, one model can compute the treatment effects of many treatments and automatically control the effect of others. We only require to expand the treatment input layer in the model $\mathcal{M}_2$ to include all the treatment variables and the causal estimate of any one of the many treatment variables can be determined by just scoring the model twice with and without that treatment variable set, and taking the difference.\\ (2) \emph{Multiple outcomes:} As CDNN inherits the flexibility and scalability of Neural Networks, it simultaneously allows for estimating the treatment effect on multiple metrics of interest by having multiple outcome variables in the output layer.\\ (3) \emph{Interaction and incremental treatment effects:} By expanding the treatment layer, CDNN can be employed to determine the impact of interaction among two or more treatment variables. By training CDNN using continuous treatment variables we could also determine treatment effect for any incremental treatment values.
\eat{
\section*{Ethical Impact}
We strongly believe that our algorithm does not put any demographic to a systematic disadvantage. In fact, our scalable framework for determining causal impact can be used across multiple domains ranging from political science to epidemiology, for applications such as estimating/forecasting the impact of a disease outbreak, impact of socio-economic events on citizens, effect of political campaign strategies on voting decisions, etc. To the best of our knowledge our paper neither violates any ethical aspect, nor presents any adverse societal consequence.
 Causal impact estimation at average or individual unit level has been extensively studied and published by researchers and widely applied  In this work, we propose a new scalable framework for efficiently computing the treatment effect at a finer, individual unit level which is much more challenging compared to obtaining average treatment effect values.}

\bibliography{cdnn}
\bibliographystyle{ACM-Reference-Format}
\newpage
\section{Supplement}
\subsection{Proof of Lemma 3.1}
\begin{proof}
Substituting the definition of $h(T,x)$ from eq.~(\ref{eq:f}) in eq.(\ref{eq:g}), we find
\begin{align}
g(x) &= e(x) \left[h(T=1,x)+g(x)\right] + \left(1-e(x)\right) \left[h(T=0, x)+ g(x)\right], \nonumber \\
\label{eq:eheq0}
\implies &e(x) \left[h(T=1,x) - h(T=0,x)\right] + h(T=0,x) = 0.
\end{align}
Plugging the identity $\theta(x) = h(T=1, x) - h(T=0, x)$ in eq.(\ref{eq:eheq0}), we get $h(T=0,x) = \theta(x) \left[0-e(x)\right]$ and $h(T=1,x) = \theta(x) \left[1-e(x)\right]$  and the proof follows.
\end{proof}

\subsection{Proof of Theorem 3.2}
\begin{proof}
Consider the score function
\begin{equation*}
\psi\left(W,\theta,\eta(x)\right) = \left(Y - g(x) - \theta(x)[T-e(x)]\right) \left(T-e(x)\right),
\end{equation*}
for which the solution $\theta = \theta_0(x)$ satisfies a system of local moment conditions
\begin{equation*}
J(\eta_0(x)) = \mathbb{E}\left[\psi\left(W,\theta_0(x),\eta_0(x)\right) | X=x\right] = 0,
\end{equation*}
where the nuisance parameter $\eta_0(x) = [g_0(x), e_0(x)]$. As stated earlier, verifying Neyman orthogonality is equivalent to establishing that the directional derivative of $J(.)$ at $\eta_0$ in the direction $\eta-\eta_0$, known as the Gateaux derivative, is zero for all $x$. Mathematically, we need to show that
\begin{equation*}
\mathbb{E}\left[\left\{\frac{\partial}{\partial \tau} \psi\left(W,\theta_0,\eta_0+\tau(\eta-\eta_0)\right)\right\}_{\tau=0} \, \middle|\, x\right] =0, \forall x,
\end{equation*}
where we have dropped the explicit dependency of $\theta$ and $\eta$ on $x$ to simplify the notation.
Let 
\begin{align*}
E(\tau) &= \psi\left(W,\theta_0,\eta_0+\tau(\eta-\eta_0)\right) \\
&= \big(Y - (1-\tau)g_0 - \tau g - \theta_0[T-(1-\tau)e_0-\tau e]\big)\big(T-(1-\tau)e_0-\tau e \big).
\end{align*}
Then, $$\frac{\partial E}{\partial \tau}_{\tau = 0} = \big(g_0 - g + \theta_0(e-e_0)\big) (T-e_0) + \big(Y-g_0-(T-e_0)\theta_0\big) (e_0-e).$$  It follows that
\begin{align*}
\mathbb{E}\left[ \left\{\frac{\partial E}{\partial \tau}\right\}_{\tau = 0} \, \middle|\, x\right] = &\big(g_0 - g + \theta_0(e-e_0)\big) \mathbb{E}_{T|x} \left[T-e_0\right] \\
&+ (e_0-e) \left(\mathbb{E}_{Y|x} \left[Y-g_0\right] - \theta_0  \mathbb{E}_{T|x} \left[T-e_0\right] \right).
\end{align*}
Recalling that $e_0(x) = \mathbb{E}\left[T \mid x\right]$ and $g_0(x) = \mathbb{E}\left[Y \mid x\right]$, we find
\begin{equation*}
\mathbb{E}\left[ \left\{\frac{\partial E}{\partial \tau}\right\}_{\tau = 0} \, \middle|\, x\right] = 0,
\end{equation*}
proving the local Neyman orthogonality.
\end{proof}

\end{document}